\documentclass[10pt,twocolumn,letterpaper]{article}

\usepackage{iccv}
\usepackage{times}
\usepackage{epsfig}
\usepackage{graphicx}
\usepackage{amsmath}
\usepackage{amssymb}
\usepackage{booktabs}
\usepackage{xcolor}
\usepackage{multirow}

\usepackage{color, colortbl}
\definecolor{mygray}{gray}{0.90}

\usepackage[pagebackref=true,breaklinks=true,letterpaper=true,colorlinks,bookmarks=false]{hyperref}

\iccvfinalcopy 

\def\iccvPaperID{11664} 

\ificcvfinal\fi

\begin{document}

\title{Order-preserving Consistency Regularization \\for Domain Adaptation and Generalization}

\author{
  Mengmeng~Jing$^{1,2}$\thanks{This work was done when Mengmeng Jing was a visiting student at University of Amsterdam.}, Xiantong Zhen$^{2}$~\thanks{Currently with United Imaging Healthcare, Co., Ltd., China.}~, Jingjing~Li$^1$, Cees~G.~M.~Snoek$^2$ \\
  $^1$University of Electronic Science and Technology of China\\
  $^2$University of Amsterdam\\
  {\tt jingmeng1992@gmail.com}  \quad {\tt zhenxt@gmail.com}  \quad  {\tt lijin117@yeah.net}\\ {\tt c.g.m.snoek@uva.nl}
}

\maketitle
\ificcvfinal\thispagestyle{empty}\fi

\begin{abstract}
Deep learning models fail on cross-domain challenges if the model is oversensitive to domain-specific attributes, \eg, lightning, background, camera angle, \etc.
To alleviate this problem, data augmentation coupled with consistency regularization are commonly adopted to make the model less sensitive to domain-specific attributes.
Consistency regularization enforces the model to output the same representation or prediction for two views of one image. 
These constraints, however, are either too strict or not order-preserving for the classification probabilities.
In this work, we propose the Order-preserving Consistency Regularization (OCR) for cross-domain tasks.
The order-preserving property for the prediction makes the model robust to task-irrelevant transformations.
As a result, the model becomes less sensitive to the domain-specific attributes.
The comprehensive experiments show that our method achieves clear advantages on five different cross-domain tasks.

\end{abstract}

\section{Introduction}
\label{sec:intro}
Deep neural networks have demonstrated their power in many computer vision tasks, especially when the training and test sets follow the same distribution. However, when we deploy a model in a real-world environment, we often encounter domain shifts between the training and test sets, which reduces the expected test-set performance and makes us unable to deploy with confidence~\cite{pan2009survey}. For some safety-critical applications, \eg, tumor recognition~\cite{guan2021domain} and autonomous driving~\cite{gong2019dlow}, a failing model is fatal.

\begin{figure}[t!]
\begin{minipage}[b]{1.0\linewidth}
  \centering
\centerline{\includegraphics[width=7.0cm]{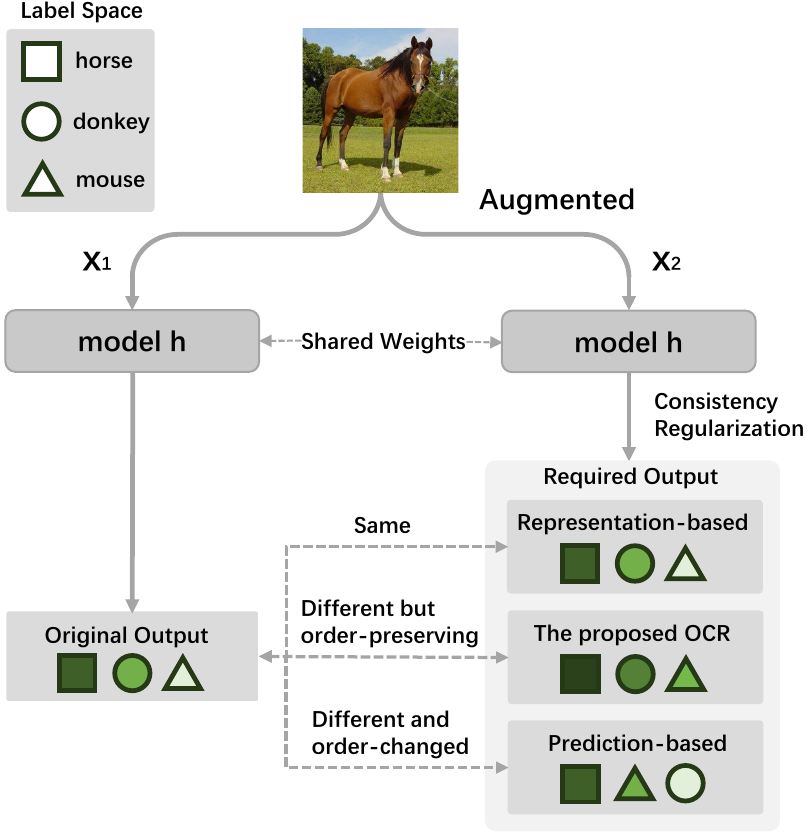}}
\end{minipage}
\caption{\textbf{The required output of three consistency regularizations.} Different shapes represent different categories. For different greens, the darker the color, the larger the classification probability. Representation-based method requires the output to be the same as the original. OCR only requires an order-preserving output and allows the output to vary. The prediction-based method is not order-preserving, which may cause the probability of the horse being classified to mouse is higher than that of donkey although donkeys are obviously more similar to horses than mice.}
\label{fig:fig1}
\vspace{-15pt}
\end{figure}

Image data consists of a variety of attributes such as shape, color, background, texture, shooting angle, \etc. We refer to one or more task-related attributes as {\it{label attributes}}, and the remaining irrelevant ones as {\it{domain-specific attributes}}.
Wiles \etal~\cite{wiles2021fine} demonstrate the domain-specific attributes cause the distribution shifts, thus weakening the generalization of the model. 
Data augmentation coupled with consistency regularization is commonly employed to make a model invariant to the domain-specific attributes~\cite{wang2022continual,chen2022contrastive,islam2021dynamic,sohn2020fixmatch,berthelot2019remixmatch,chen2020simple,chen2021exploring,grill2020bootstrap}. Data augmentation perturbs the data so that the domain-specific information is incorporated into the perturbed image. By imposing a consistency regularization on the representations of the same image before and after perturbation, the model becomes less sensitive to the domain-specific attributes. 

The existing consistency regularization methods can be divided into two categories: representation-based methods~\cite{laine2017temporal,tarvainen2017mean,sajjadi2016regularization} and prediction-based methods~\cite{berthelot2019remixmatch,miyato2018virtual,xie2020unsupervised}. For the representation-based methods, usually the $\ell_1$ or $\ell_2$ loss is employed to enforce the model to output the same representation, even though two different views are fed into the model.
This constraint, however, is too strict, which may bring difficulties to the training of the model. 
For example, different works on self-supervised learning~\cite{chen2020simple,chen2021exploring,grill2020bootstrap} have reached a consensus that one of the representations needs to go through a non-linear prediction head before performing consistency regularization with the other.
With the network model being a symmetric structure, directly imposing consistency regularization on the two representations will result in a model collapse.

Alternatively, the prediction-based methods~\cite{berthelot2019remixmatch,miyato2018virtual,xie2020unsupervised} employ the cross-entropy loss to regularize the maximum classification probability of two representations to be the same. In other words, they ignore the order of the other classes, which would reduce the discriminability of the model. 
For example, consider a classification problem of three classes: {\it horse, donkey and mouse}. As illustrated in Fig.~\ref{fig:fig1}, for an image of a horse, the cross-entropy loss only regularizes the maximum classification probability of two representations to be horse, but it ignores the classification probability of donkey and mouse. If the order of classification probability is horse$>$donkey$>$mouse before augmentation, it may become horse$>$mouse$>$donkey after augmentation. Although the classification results have not changed, the discrimination of the model has reduced as donkeys are obviously more similar to horses than mice.

In view of these problems, we propose Order-preserving Consistency Regularization (OCR) for cross-domain tasks.
OCR is able to enhance the model robustness to domain-specific attributes without the need of an asymmetric achitecture or a stop gradient. Specifically, we compute the residual component which is the variation in the augmented representation relative to the original representation. We postulate that if the model is robust to domain-specific attributes, the residual component should contain little or no task-related information. For example, in the classification task, when we classify the residual component, we regularize it to have the same probability to be classified into each category. In this way, the classification probabilities of the augmented representation are order-preserving compared to the original representation. As a result, the model becomes less sensitive to the domain-specific attributes. The core idea of OCR is that we allow the model to output different representations for two views of the same image, as long as the residual component contains as little task-related information as possible.

The contributions of this paper are threefold:
\begin{enumerate}
    \item We propose Order-preserving Consistency Regularization (OCR) to enhance model robustness to domain-specific attributes. Compared with representation-based methods, OCR relaxes the constraints on model training, \ie, it allows the model to output different representations for two views of the same image. Compared with prediction-based method, OCR maintains the order of the classification probabilities before and after augmentation, which helps the model to be less sensitive to the domain-specific attributes.
    \item We provide a theoretical analysis for OCR. We prove that the representation-based method is a special case of OCR. Moreover, OCR can reduce the mutual information between the domain-specific attributes and the label attribute.
    \item We test our method on five different cross-domain vision tasks to demonstrate the effectiveness of our method. In particular, OCR helps to enhance the robustness of the model against adversarial attacks.
\end{enumerate}

\section{Related Work}

\noindent{\bf Consistency Regularization.}
Consistency regularization~\cite{bachman2014learning,laine2017temporal,sajjadi2016regularization} is a common self-supervised learning method which enforces the model to output the same prediction even when the input is perturbed. Since it can enhance the robustness of the model to domain-variant styles, it has recently been used to address cross domain challenges~\cite{chen2022contrastive,wang2022continual}. To generate the perturbed version of the image, some methods employ adversarial training~\cite{miyato2018virtual} or dropout~\cite{laine2017temporal,tarvainen2017mean}, while others add perturbations by applying heuristic data augmentations~\cite{laine2017temporal,sajjadi2016regularization,berthelot2019mixmatch,xie2020unsupervised}, such as color jitter, Gaussian blur, rotate, cutout, \etc. To measure the consistency, the $\ell_1$ or $\ell_2$ norm~\cite{laine2017temporal,tarvainen2017mean,sajjadi2016regularization} are adopted.
Given the images of the original version and the perturbed version, some methods~\cite{xie2020unsupervised,sajjadi2016regularization,chen2022contrastive} employ the same model to extract representations for the two images, and then impose consistency regularization. We believe that this strategy is too strict, thereby increasing the difficulty of model training.

\begin{figure*}[t]
\centering
\centerline{\includegraphics[width=16cm]{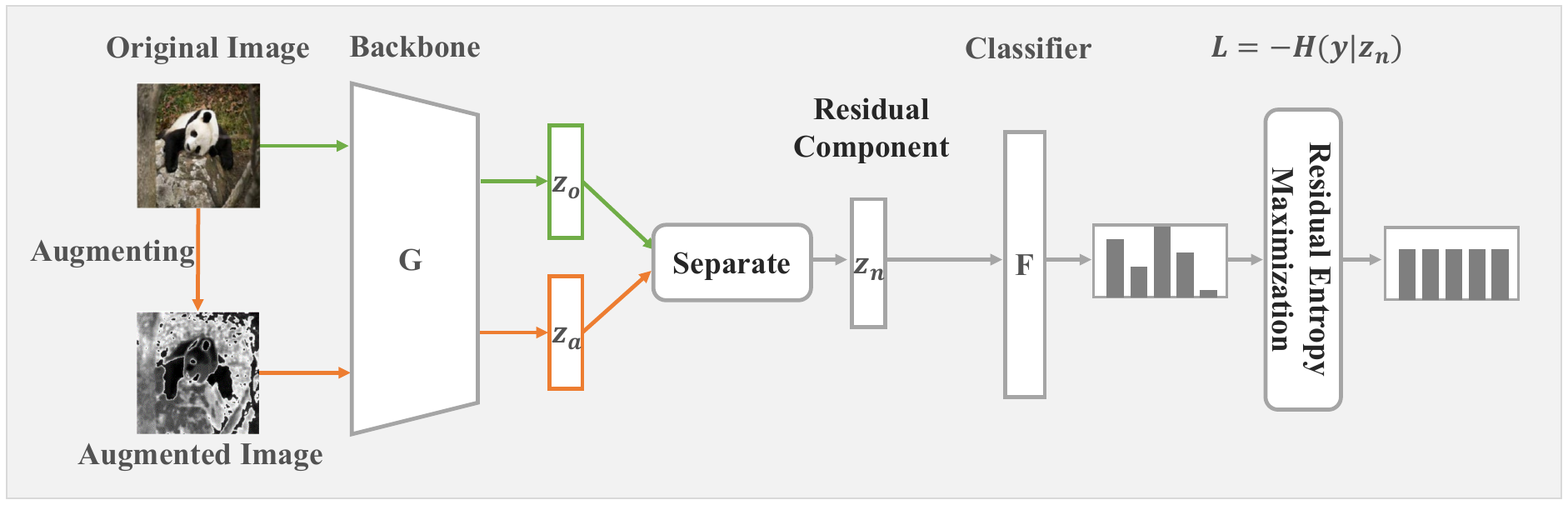}}
\caption{\normalsize{\textbf{Method overview}. Given the original image and its augmented counterpart, we feed them into the backbone model to obtain the original representation $z_o$ and the augmented representation $z_a$. Then, we compute their residual component $z_n$ and feed it into the classifier to get the classification results. Finally, we maximize the entropy of the classification probabilities for the residual component to reduce the task-related information in the residual component. As a result, the model become less sensitive to the domain-specific attributes.}}
\label{fig:overview}
\end{figure*}

To solve this problem, some works~\cite{chen2020simple,chen2021exploring,grill2020bootstrap} have designed an asymmetric architecture for the model, where one representation needs to go through an additional non-linear layer, which makes two images go through different paths in the same model. Although effective, these methods increase the complexity of the model architecture, and often require a large number of training data to unleash their power. Another line of work feeds one of the images (usually the original version) into the running average model or past model and then applies consistency regularization~\cite{laine2017temporal,tarvainen2017mean}. These methods allow two versions of the images to pass through two similar yet different models, alleviating the problem of overly strict regularization to some extent. However, these methods require the storage of multiple copies of the model, thereby increasing the GPU memory consumption. 
Different from the above methods, our method does not require an asymmetric architecture, nor does it need to store multiple copies of the models. Our method allows the model to output different representations for two versions of the images, as long as the residual component obtained from these two representations does not contain task-related information.

\noindent{\bf Domain Adaptation and Generalization.}
Both domain adaptation (DA)~\cite{long2018conditional,saito2018maximum,liang2020we,ganin2015unsupervised,tzeng2017adversarial,zhang2018collaborative} and domain generalization (DG)~\cite{muandet2013domain,ghifary2015domain,niu2015visual,motiian2017unified,niu2015multi} are cross-domain tasks, but their task settings are different.
In the DA task, we are given a labeled source domain and an unlabeled target domain. We use the joint training of source and target samples to make the model adapt to the shifts between domains. The recent focus on privacy and copyright has given rise to a variant of the vanilla DA, \ie, source-free domain adaption (SFDA)~\cite{liang2020we,yang2021exploiting,huang2021model,jing2022variational}, where we are given a pre-trained source model but cannot directly access the source data. Based on SFDA, a new setting is proposed, namely Test-Time Adaptation (TTA)~\cite{wang2020tent,wang2022continual,chen2022contrastive,xiao2021learning}. TTA further requires that the model can adapt to the target domain in an online fashion, which is an even more challenging setting.

DG~\cite{muandet2013domain,ghifary2015domain,niu2015visual,motiian2017unified,niu2015multi} trains on one or more labeled source domains to learn a model that is robust to changes in domain shifts, so that the trained model generalizes well to the (unseen) target domain. Compared to DA, DG is more difficult because during training it does not have access to (unlabeled) data from the target domain to adapt to changes in the distribution. The commonality between DA and DG is that they strive to learn the domain-invariant representations for better performance on the target domain. 
OCR can regularize the order of the predictions so that the model is insensitive to the domain-specific attributes, thus alleviating the domain shifts.

\section{Methodology}
\textbf{Problem Formulation.}
In many computer vision challenges, be it image classification or semantic segmentation, we are given a dataset $\mathcal{D}_{train}$ \!\!{=}\{x$\in\!\!\mathcal{X}_{train}$,y$\in\mathcal{Y}_{train}$\} where $\mathcal{X}_{train}$ and $\mathcal{Y}_{train}$ are the image set and label set for training and we need to establish the relationship between the data $\mathcal{X}_{train}$ and the ground-truth label $\mathcal{Y}_{train}$. In the classical Empirical Risk Minimization (ERM)~\cite{Vapnik1998}, the training objective is to choose a hypothesis $h:\mathcal{X}\to\mathcal{Y}$ from a pre-defined hypothesis space $\mathcal{H}$ where the empirical risk is minimized w.r.t $\mathcal{D}_{train}$: inf$_{h\in \mathcal{H}} {=} \mathbb{E}_{(x,y)\sim \mathcal{D}_{train}}$[$\mathcal{L}(h(x),y)$].
However, when deployed to the test set $\mathcal{D}_{test}$, the model would suffer from performance degradation since there may be domain shifts between the training set $\mathcal{D}_{train}$ and the test set $\mathcal{D}_{test}$, \ie, $P(\mathcal{X}_{train})\neq P(\mathcal{X}_{test})$. 
For example, samples of the same category in the training set and the test set often have varying appearance, caused by lightning conditions, camera angle, background, \etc. These accidental attributes are irrelevant to our task, but will cause domain shifts. To generalize well, we need to train the model to be invariant to these domain-specific attributes. 

\textbf{Order-preserving Consistency Regularization.} For a global understanding, we provide the overview of our method in Fig.~\ref{fig:overview}. OCR consists of three steps, \ie, data augmentation, residual component separation, and residual entropy maximization.
Data augmentation~\cite{bachman2014learning,sajjadi2016regularization,laine2017temporal,miyato2018virtual,xie2020unsupervised} is a commonly used technology, which increases the diversity of samples and helps to improve the generalization of the model.
Given a sample $x_{o}$, we obtain its augmentated version $x_{a} {=} \mathcal{N}(x_{o})$ using transformations $\mathcal{N}$.
For a clearer narration, we split the hypothesis $h$ into two parts, \ie, $h {=} F \circ G$, where $G$ is the backbone model and $F$ is the classifier. We feed $x_{o}$ and $x_{a}$ into $G$ to get two different representations of the same sample: $z_{o} {=} G(x_{o})$, $z_{a} {=} G(\mathcal{N}(x_{o}))$.

We define the residual component as the variation in the augmented representation relative to the original representation. To separate the residual component, an intuitive method is to subtract the original representation from the augmented representation. In order to control the proportion of the residual component more flexibly, here we consider the following linear relations:
\begin{equation}
    z_{a} = \lambda  z_{o} + (1 - \lambda) z_n, \label{eq:linear_comb}
\end{equation}
where $z_n$ is the residual component, $\lambda \in (0,1)$ represents the proportion of $z_{o}$ in $z_{a}$. 
From the perspective of Occam’s razor, linearity is a good inductive bias, as also used in mixup~\cite{zhang2018mixup}.
Another reason we choose the relation in Eq. (\ref{eq:linear_comb}) is that it is an invertible operation so that we can easily infer $z_n$ given $z_{a}$ and $z_{o}$:
\begin{equation}
    z_n = \frac{z_{a} - \lambda z_{o}}{ 1 - \lambda}. \label{eq:zn}
\end{equation}

With the residual component $z_n$, we try to maximize the uncertainty of $z_n$'s prediction so that it does not contain too much classification-related information. As the entropy can be regarded as the measure of the prediction uncertainty, we maximize the conditional entropy ${\mathcal{H}}(y|z_n)$ to enlarge the uncertainty of $z_n$'s prediction. Therefore, our objective is as follows:
\begin{equation}
    \mathcal{L}_{\rm OCR} = -{\mathcal{H}}(y|z_n) = -{\mathcal{H}}[ {\mathrm {Softmax}}(F(z_n))] , \label{eq:overall}
\end{equation}
where $F(z_n)\in {\mathbb R}^{B\times C}$ is the prediction of $z_n$, $B$ is the batch size, $C$ is the category number. 
$\mathcal{H}$ is the entropy.
By minimizing Eq. (\ref{eq:overall}), we regularize $z_n$ to have equal probability of being classified into each category.

During training, we use $\lambda$ to control the proportion of the residual component and the original representation in the augmented representation.
$\lambda$ should change dynamically to match the process of model training.
At the beginning of training, the model would be sensitive to the domain-specific attributes, so the difference between $z_{o}$ and $z_{a}$ would be large. Then, $\lambda$ should be a small value so that the proportion of $z_{o}$ in $z_{a}$ is lower.
As the training goes on, the model gradually becomes less sensitive to the domain-specific attributes, at this time, $z_{o}$ and $z_{a}$ would be similar and $\lambda$ should increase to a larger value accordingly.
Inspired by Ganin \etal~\cite{ganin2016domain}, we adopt an annealing strategy for $\lambda$:
\begin{equation}
    \lambda = \lambda_0[1-(1+\alpha \frac{t}{T})^{-\beta}], \label{eq:lambda}
\end{equation}
where $\alpha{=}10$, $\beta{=}0.75$, $t$ is the current iteration number and $T$ is the total number of iterations. $\lambda_0$ is the initial value of $\lambda$.
In this way, $\lambda$ is more likely to be sampled to a small value at the beginning of training and then gradually becomes larger as the training goes on. In the ablations we illustrate that this strategy could achieve better performance than that of a fixed $\lambda$ value.

\begin{table*}[t]
\centering
\caption{\textbf{Overview} of tasks, datasets, backbones and evaluation metrics.}
\label{tab:statics_data}
\vspace{5pt}
\resizebox{.9\textwidth}{!}{%
\begin{tabular}{llll}
\hline
Task                                 & Dataset        & Backbone    & Evaluation metric           \\ \hline
\textbf{Domain Adapatation Classification}                   & Office-Home    & ResNet-50     & Accuracy         \\ 
\textbf{Test-Time Adaptation }                & CIFAR100-C     & ResNeXt-29     &  Accuracy      \\ 
\textbf{Domain Generalization Classification} & PACS           & ResNet-18    & Accuracy      \\ 
\multirow{2}{*}{\textbf{Domain Generalization Segmentation}}   & GTAV, SYNTHIA, Cityscapes & \multirow{2}{*}{DeepLabV3+}  & \multirow{2}{*}{mIoU} \\
                                                      & BDD100K, Mapillary  & &\\
                                                      \textbf{Domain Adaptation Object Detection} & Cityscapes, FoggyCityscapes & ResNet-50 & mAP \\\hline
\end{tabular}%
}
\end{table*}

Now we prove three properties of the proposed OCR:

{\bf (1) OCR is order-preserving.} In previous methods with consistency regularization, \eg,~\cite{chen2022contrastive,sajjadi2016regularization,xie2020unsupervised}, the similarity between the representations and the prototype feature of a class in classifier $F$ is computed as:
\begin{equation}
    \hat{y}_{o}^i = \textrm{sim}(P_i, z_{o}), \hat{y}_{a}^i =  \textrm{sim}(P_i, z_{a}), \label{eq:yaug}
\end{equation}
where $\textrm{sim}(\cdot)$ is the similarity function, \ie, the inner-product, $P_i$ is the prototype feature of class $i$, $\hat{y}_{o}^i$ and $\hat{y}_{a}^i$ are probabilities of representations $z_{o}$ and $z_{a}$ belonging to class $i$, respectively.
When substituting Eq. (\ref{eq:linear_comb}) into Eq. (\ref{eq:yaug}), we get:
\begin{align}
  \hat{y}_{a}^i &= \textrm{sim}(P_i, \lambda z_o + (1-\lambda)z_n) \notag\\ &= \lambda \textrm{sim}(P_i, z_{o}) + (1-\lambda) \textrm{sim}(P_i, z_{n}) \notag\\ &= \lambda \hat{y}_{o}^i + (1-\lambda) \hat{y}_{n}^i.
\end{align}
In Eq. (\ref{eq:overall}), when the conditional entropy ${\mathcal{H}}(y|z_n)$ is maximized, the residual component will have equal probability of being classified into each category, \ie, $\hat{y}_{n}^1 = \hat{y}_{n}^2=\cdots=\hat{y}_{n}^C=K$. Therefore, the relation between $\hat{y}_{a}^i$ and $\hat{y}_{o}^i$ is:
\begin{equation}
    \hat{y}_{a}^i = \lambda \hat{y}_{o}^i + (1-\lambda) K = f(\hat{y}_{o}^i;\lambda,K).
\end{equation}
Within the same iteration, $K$ and $\lambda$ are two constants. Then, $f(\hat{y}_{o}^i;\lambda,K)$ is an order-preserving mapping, which guarantees that if $\hat{y}_{o}^j>\hat{y}_{o}^k$, then $\hat{y}_{a}^j>\hat{y}_{a}^k$. 
Therefore, OCR is order-preserving.

{\bf (2) Representation-based consistency regularization is a special case of OCR}. 
Previous cross-domain methods, \eg, \cite{xie2020unsupervised,sajjadi2016regularization,chen2022contrastive}, optimize the $\ell_1$ or $\ell_2$ loss to impose consistency regularization between $z_o$ and $z_a$. 
We use $\hat{z}_n {=} z_{a} - z_{o}$ to represent the unnormalized residual component, $\Delta y^i$ to represent the prediction of $\hat{z}_n$ belonging to class $i$.
Therefore, the objective of representation-based consistency regularization is to make $\hat{z}_n$ close to the zero vector:
\begin{equation}
    \hat{z}_n = {\bf 0} \Rightarrow \textrm{sim}(P_i, \hat{z}_n) = 0 \Rightarrow \Delta y^1  = \cdots = \Delta y^C = 0. \label{cons_reg_obj}
\end{equation}
We believe this regularization is too strict and may increase the training difficulty. It is very reasonable for the model to output different representations for different images. 
The goal of our method is {\it not} to enforce $\hat{z}_n$ to be close to the zero vector, but to make $\hat{z}_n$ contain no task-related information. Our method relaxes the constraint in Eq. (\ref{cons_reg_obj}) as:
\begin{equation}
    \Delta y^1 = \Delta y^2 = \cdots = \Delta y^C. \label{our_obj}
\end{equation}
Obviously, $\hat{z}_n$ of the zero vector in Eq. (\ref{cons_reg_obj}) can also match the condition in Eq. (\ref{our_obj}), making representation-based consistency regularization a special case of our method.

{\bf (3) OCR can make the model less sensitive to the domain-specific attributes}. 
The mutual information between the residual component and the label attribute is:
\begin{align}
    I(Z_n; Y) = & KL[ p(z_n, y) || p(z_n)p(y)   ]\\
                = & \int d z_n\, d y \,p(z_n, y) {\rm log}\frac{p(z_n, y)}{p(z_n)p(y)}\\
                = & \int d z_n\, d y\, p(z_n, y) {\rm log}\frac{p(y|z_n)}{p(y)}\\
                = & \mathcal{H}(Y) - \mathcal{H}(Y|Z_n),
\end{align}
where $\mathcal{H}$ denotes the entropy, $Y$ is the label set and $Z_n$ is the residual component set, $z_n\in Z_n$, $y\in Y$. Note that $\mathcal{H}(Y)$ is independent of our optimization procedure and so can be ignored. Then, we have:
\begin{equation}
   \!\! \mathop{\min}\limits_{z_n} I(Z_n; Y) = \mathop{\min}\limits_{z_n} - \mathcal{H}(Y|Z_n) = \mathop{\max}\limits_{z_n} \mathcal{H}(Y|Z_n). \label{theorem_obj}
\end{equation}  
Therefore, by minimizing Eq.~(\ref{eq:overall}), we are just minimizing the mutual information between the residual component and the label attribute. 
As data augmentation imposes various task-irrelevant transformations to introduce domain-specific attributes for the original representation and correspondingly generates the residual component, the residual component can be regarded as the proxy of domain-specific attributes.
Minimizing the mutual information in Eq. (\ref{theorem_obj}) can decorrelate the domain-specific attributes and the label attribute.
As a result, the problem of sensitivity to domain-specific attributes is alleviated. 

According to the Information Bottleneck principle~\cite{tishby1999information}, an optimal representation $z$ of input $x$ should satisfy two properties: sufficiency and minimality. 
Achille and Soato \cite{achille2018emergence} have demonstrated that being invariant to domain-specific attributes is helpful to guarantee the minimality. Therefore, the proposed OCR is helpful to learn a better representation, which could improve the performance of the model on cross-domain tasks.

\section{Experiments}

\subsection{Tasks, Datasets and Setup} 
To evaluate our method, we consider five different cross-domain tasks: domain adaptation, test-time adapatation, domain generalization classification, domain generalization detection, and domain generalization semantic segmentation. Different tasks involve different datasets and setups, which we summarize in Table~\ref{tab:statics_data}.

\textbf{Domain Adaptation Classification.} For domain adaptation classification we report on \textit{Office-Home}~\cite{venkateswara2017deep}. It consists of four domains: Art, Clipart, Product and Real-world. There are about 15,500 images categorized into 65 classes. 
We consider two different settings, \ie, source-dependent ~\cite{long2018conditional,saito2018maximum} and source-free~\cite{liang2020we,yang2021exploiting}. For the source-dependent setting, we use all labeled source samples and all unlabeled target samples for training. For the source-free setting, only the model trained in the source domain and the unlabeled target samples are given. Upon evaluation, we test the models in the unlabeled target samples. For the hyper-parameter, we set $\lambda_0{=}0.7$.

\textbf{Test-Time Adaptation.}  For test-time adaptation we report on \textit{CIFAR100-C}~\cite{hendrycks2018benchmarking}. This dataset includes 15 different corruption types with five levels of severity categorized into 100 classes.  These corruptions were added to clean images from CIFAR100~\cite{krizhevsky2009learning}. There are 10,000 images for each corruption type. We used the ResNeXt-29 model pre-trained in the clean CIFAR100 dataset from \cite{hendrycks2020augmix}. This task involves two settings: online~\cite{wang2020tent} and continual online~\cite{wang2022continual}. In both settings, we conduct the experiments on CIFAR100-C in an online fashion without the need for labels. The difference between the two settings is that the online setting will initialize the model to the state of pre-training on the clean dataset before adapting to each corruption type, while the continual online setting will continuously adapt data of different corruption types. In this task, we evaluate our method on images with the highest severity, \ie, level 5.
For the hyper-parameter, we set $\lambda_0{=}0.8$.

\textbf{Domain Generalization Classification.} For this task we report on \textit{PACS}~\cite{li2017deeper}. A commonly used domain generalization benchmark which includes four domains: Art Painting, Cartoon, Photo and Sketch. There are 9,991 images categorized into seven classes. We train the model on 3 of 4 domains and evaluate it on the remaining one. In this task, we set $\lambda_0{=}0.5$. 

\begin{table}[t]
\small
\centering
\caption{\textbf{Domain Adaptation.} Accuracy (\%) on Office-Home with ResNet-50 backbone. All per-domain results are in the supplementary material. R- and P-Cons. Reg. means representation-based and prediction-based consistency regulariztion. Results with  are implemented by us.}
\label{tab:home}
\vspace{5pt}
\resizebox{.48\columnwidth}{!}{%
\begin{tabular}{lc}
\toprule
Method  & Mean \\ 
\midrule
\rowcolor{mygray}
\textbf{Source-Use} &\\
MCD~\cite{saito2018maximum}  & 64.1 \\
\quad {\bf w/ OCR}~\cite{saito2018maximum}   & 66.6 \\
CDAN~\cite{long2018conditional}      & 65.8 \\
\quad {\bf w/ OCR}  & 68.0 \\
\rowcolor{mygray}
\textbf{Source-Free} &\\
ResNet-50~\cite{he2016deep} & 46.1 \\
Source-only  & 60.2 \\   
NRC~\cite{yang2021exploiting}  &  71.9 \\ 
\quad w/ R-Cons. Reg.  &  71.5 \\
\quad w/ P-Cons. Reg.  &  72.1 \\
\quad {\bf w/ OCR} &  {\bf 72.6} \\
SHOT~\cite{liang2020we}  & 71.8 \\
\quad w/ R-Cons. Reg. &  71.4 \\ 
\quad w/ P-Cons. Reg. &  72.0 \\
\quad {\bf w/ OCR}  &  {\bf 72.8} \\ 
SHOT++~\cite{liang2021source}  & 72.8 \\
\quad {\bf w/ OCR}  &  {\bf 73.2} \\
\bottomrule
\end{tabular}%
}
\end{table}

\begin{table}[t]
\centering
\caption{\textbf{Test-Time Adaptation.} Test error (\%) for CIFAR100-to-CIFAR100C adaptation. The backbone model is ResNeXt-29. The corruption severity is 5. OCR can improve the baselines on both online setting and continual online setting.}
\label{tab:cifar100c}
\vspace{5pt}
\resizebox{0.38\textwidth}{!}{%
\begin{tabular}{lcc}
\toprule
        & Online & Continual Online \\\midrule
Source        &  46.4  & 46.4             \\
BN Adapt~\cite{li2017revisiting}      &  35.8  & 35.4             \\\hline
TENT~\cite{wang2020tent}          & 34.4   & 35.6             \\
\quad {\bf w/ OCR}  & {\bf 31.3}   & {\bf 32.4}             \\\hline
CoTTA~\cite{wang2022continual}         & 36.8   & 32.5             \\
\quad {\bf w/ OCR} & {\bf 34.6}   & {\bf 31.6}             \\\bottomrule  
\end{tabular}%
}
\vspace{-5pt}
\end{table}

\begin{table}[t]
\Huge
\centering
\caption{\textbf{Domain Generalization Classification.} Accuracies (\%) on PACS. Results are based on the leave-one-domain-out protocol \cite{zhou2020domain}, where for each task we use 3 of the 4 domains as the source and the remaining 1 as the target, \eg, "Art" means "Cartoon, Photo, Sketch$\to$Art". R- and P-Cons. Reg. means representation-based and prediction-based consistency regulariztion.}
\label{tab:dgc}
\vspace{5pt}
\resizebox{0.48\textwidth}{!}{%
\begin{tabular}{lccccc} 
\toprule
& \multicolumn{4}{c}{\textbf{PACS}} & \\
	\cmidrule(lr){2-5}
          & Art      & Cartoon  & Photo    & Sketch   & Mean \\ \midrule
MMD-AAE~\cite{li2018domain}         & 75.2     & 72.7     & 96.0     & 64.2     & 77.0 \\
CCSA~\cite{motiian2017unified}            & 80.5     & 76.9     & 93.6     & 66.8     & 79.4 \\
JiGen~\cite{carlucci2019domain}           & 79.4     & 75.3     & 96.0     & 71.6     & 80.5 \\
Metareg~\cite{balaji2018metareg}         & 83.7     & 77.2     & 95.5     & 70.3     & 81.7 \\
L2A-OT~\cite{zhou2020learning}          & 83.3     & 78.2     & 96.2     & 73.6     & 82.8 \\\hline
ResNet-18~\cite{he2016deep}       & 77.5 & 77.9 & 96.1 & 70.7 & 80.6 \\
\quad { w/ Manifold Mixup}~\cite{verma2019manifold} & 75.6 & 70.1 & 93.5 & 65.4 & 76.2 \\
\quad { w/ Cutout}~\cite{devries2017improved}         & 74.9 & 74.9 & 95.9 & 67.7 & 78.3 \\
\quad { w/ Cutmix}~\cite{yun2019cutmix}         & 74.6 & 71.8 & 95.6 & 65.3 & 76.8 \\
\quad { w/ Mixup}~\cite{zhang2018mixup}          & 76.8 & 74.9 & 95.8 & 66.6 & 78.5 \\
\quad { w/ DropBlock}~\cite{ghiasi2018dropblock}      & 76.4 & 75.4 & 95.9 & 69.0 & 79.2 \\
\quad { w/ MixStyle}~\cite{zhou2020domain}       & 82.3 & 79.0 & {\bf 96.3} & 73.8 & 82.8 \\
\quad { w/ R-Cons. Reg.}        & 77.9 & 78.6 & 93.5 & 78.6 & 82.2     \\
\quad { w/ P-Cons. Reg.}        & 79.2 & 80.2 & 95.9 & 79.3 & 83.7     \\
\quad {\bf \,w/ OCR}        & 84.4     & 80.7     & 95.9     & 80.8     & {\bf 85.5} \\ \midrule
IIB~\cite{li2022invariant} & 79.5     & 80.3     & 96.0     & 79.8     & 83.9 \\ 
\quad {\bf \,w/ OCR}        & 85.1     & 80.9     & 96.2     & 81.8     & {\bf 86.0} \\ 
SFA~\cite{li2021simple} & 81.2     & 77.8     & 93.9     & 73.7     & 81.7 \\ 
\quad {\bf \,w/ OCR}        & 84.5     & 80.5     & 96.1     & 81.2     & {\bf 85.6} \\ 
SelfReg~\cite{kim2021selfreg} & 82.3     & 78.4     & 96.2     & 77.5     & 83.6 \\ 
\quad {\bf \,w/ OCR}        & 85.5     & 80.9     & 96.2     & 81.4     & {\bf 86.0} \\ 
CIRL~\cite{lv2022causality} & 86.1 & 81.0 & 95.9 & {\bf 82.7} & 86.3 \\
\quad {\bf \,w/ OCR}        & {\bf 86.3}     & {\bf 81.5}     & 96.1     & 82.4     & {\bf 86.6} \\ 
\bottomrule
\end{tabular}%
}
\vspace{-5pt}
\end{table}

\begin{table*}[t]
\centering
\caption{ \textbf{Domain Generalization Semantic Segmentation.} 
All models are trained on GTAV and evaluated on BDD100K, Cityscapes, SYNTHIA and Mapillary. We use ResNet-50 with output stride 16. Results with $\dag$ are from \cite{choi2021robustnet}. Best mIoU (\%) results highlighted in bold. OCR can improve all the baseline methods.}
\label{tab:segment}
\vspace{5pt}
\begin{tabular}{lcccccc}
\toprule
Source            & \multicolumn{4}{c}{GTAV}                   & \multirow{2}{*}{Mean} & \multirow{2}{*}{Boost} \\\cline{1-5}
Target            & BDD100K & Cityscapes & SYNTHIA & Mapillary &                       &                        \\\midrule
$\dag$DeepLabv3+~\cite{chen2018encoder}          & 25.1    & 29.0       & 26.2    & 28.2      & 27.1                  &                        \\
\quad {\bf w/ OCR}  & 34.7    & 34.8       & 25.1    & 39.8      & 33.6                  & 6.5   $\uparrow$                 \\\hline 
$\dag$IBN-Net~\cite{pan2018two}           & 32.3    & 33.9       & 27.9    & 37.8      & 33.0                  &                        \\
\quad {\bf w/ OCR}   & 34.9    & {\bf 41.7}       & 27.6    & 38.7      & {\bf 35.7}                  & 2.7 $\uparrow$                   \\\hline 
$\dag$RobustNet~\cite{choi2021robustnet}         & 35.2    & 36.6       & {\bf 28.3}    & {\bf 40.3}      & 35.1                  &                        \\
\quad {\bf w/ OCR} & {\bf 37.2}    & 38.9       & 27.0    & 39.7      & {\bf 35.7}                  & 0.6 $\uparrow$  \\\bottomrule                
\end{tabular}%
\end{table*}

\begin{table}[t]
\centering
\caption{ \textbf{Domain Adaptation Object Detection.} mAP (\%) on Cityscapes $\to$ FoggyCityscapes. }
\label{tab:od}
\vspace{5pt}
\begin{tabular}{lcc}
\toprule
Method & mAP & Boost  \\ \midrule
SDAT~\cite{rangwani2022closer}   & 37.5 & \\
\quad {\bf w/ OCR} & {\bf 39.1} & 1.6 $\uparrow$ \\ \midrule
SUDA~\cite{zhang2022spectral}   & 42.8 &  \\
\quad {\bf w/ OCR} & {\bf 44.2} & 1.4 $\uparrow$ \\ \bottomrule
\end{tabular}
\end{table}

\textbf{Domain Generalization Semantic Segmentation.} For this task we follow the \textit{Semantic segmentation benchmark}~\cite{choi2021robustnet}, which includes five datasets. \textit{GTAV}~\cite{richter2016playing} is a large-scale synthetic dataset consisting of 24,966 driving-scene images generated from the Grand Theft Auto V game. There are 19 objects in the images. \textit{SYNTHIA}~\cite{ros2016synthia} is another synthetic dataset containing 9,400 photo-realistic synthetic images with a resolution of 960$\times$720.
\textit{Cityscapes}~\cite{cordts2016cityscapes} is a large-scale real-world dataset consisting of 3,450 finely-annotated images and 20,000 coarsely-annotated images collected from urban scenes of 50 cities in Germany. We use the finely-annotated set for training and testing. \textit{BDD-100K}~\cite{yu2020bdd100k} is also a real-world dataset which consists of urban driving scene images collected from the US. We use 7,000 images for training and 1,000 images for evaluation. \textit{Mapillary}~\cite{neuhold2017mapillary} is the last real-world dataset containing 25,000 images collected from locations around the world.
For this task, we follow the protocol in \cite{choi2021robustnet}. Specifically, the model is trained in GTAV for 40K iterations and evaluated on the remaining datasets. In this task, we set $\lambda_0{=}0.1$.

\textbf{Domain Adaptation Object Detection.} In this task, we report on \textit{Cityscapes}~\cite{cordts2016cityscapes} and \textit{FoggyCityscapes}~\cite{sakaridis2018semantic}. FoggyCityscapes~\cite{sakaridis2018semantic} is a synthetic foggy dataset based on Cityscapes. Each image is rendered with a clear Cityscapes image and the depth map.
There are 8 categories in both domains. For the hyper-parameter, we set $\lambda=0.5$.

For the data augmentations used in our method, we apply {\tt RandomCrop} and {\tt RandomHorizontalFlip} for the original image. For the augmented images, we further apply {\tt ColorJitter}, {\tt RandomGrayscale} and {\tt GaussianBlur}. The detailed parameters for these augmentations are in the supplementary material.

To test the effectiveness of our method, in all the cross-domain tasks, our method is inserted into the existing method as a plug-and-play module. We choose the weight of OCR through importance-weighted cross validation~\cite{sugiyama2007covariate}. Our method is implemented with PyTorch~\cite{paszke2019pytorch} and MindSpore\footnote{\url{https://www.mindspore.cn}}. Code is available at \url{https://github.com/mmjing/OCR}.

\subsection{Results}

\textbf{Domain Adaptation.} 
In Table~\ref{tab:home}, OCR is inserted as a plug-and-play module in each of the compared methods. 
In the source-dependent setting, OCR improves MCD~\cite{saito2018maximum} by 2.5\% and CDAN~\cite{long2018conditional} by 2.2\%.
In the source-free setting, OCR is still effective, it improves NRC~\cite{yang2021exploiting} by 0.7\% and SHOT~\cite{liang2020we} by 1.0\%.
In addition, as a comparison, OCR outperforms the prediction-based consistency regularization. We observe that the representation-based method does not offer clear advantage over the baseline NRC~\cite{yang2021exploiting} and SHOT~\cite{liang2020we}, which may be due to the strict regularization that increases the difficulty of model training.
OCR is feature-based and independent of specific architectures, so it can be applied to transformer-based methods as well. We test SDAT~\cite{rangwani2022closer} (ViT-B/16) and SDAT with OCR on Office-Home and achieve results of 84.3\% and 85.0\%, respectively. OCR can also achieve performance improvements on transformer-based architectures.

\textbf{Test-Time Adaptation.} 
In Table~\ref{tab:cifar100c}, for the online setting, OCR achieves 3.1\% and 2.2\% lower test errors than TENT~\cite{wang2020tent} and CoTTA~\cite{wang2022continual}, respectively. For the continual online setting, TENT~\cite{wang2020tent} and CoTTA~\cite{wang2022continual} are also improved by 3.2\% and 0.9\% after adding OCR.
This shows that OCR can enhance the robustness of the model against various types of corruptions. In fact, the augmented data can be regarded as data with corruption applied. Our OCR can effectively reduce the task-related information residing in the residual component in the augmented representations, thus enhancing the robustness of the model.

\begin{figure}[t]
\begin{minipage}[b]{0.50\linewidth}
  \centering
\centerline{\includegraphics[width=4.0cm]{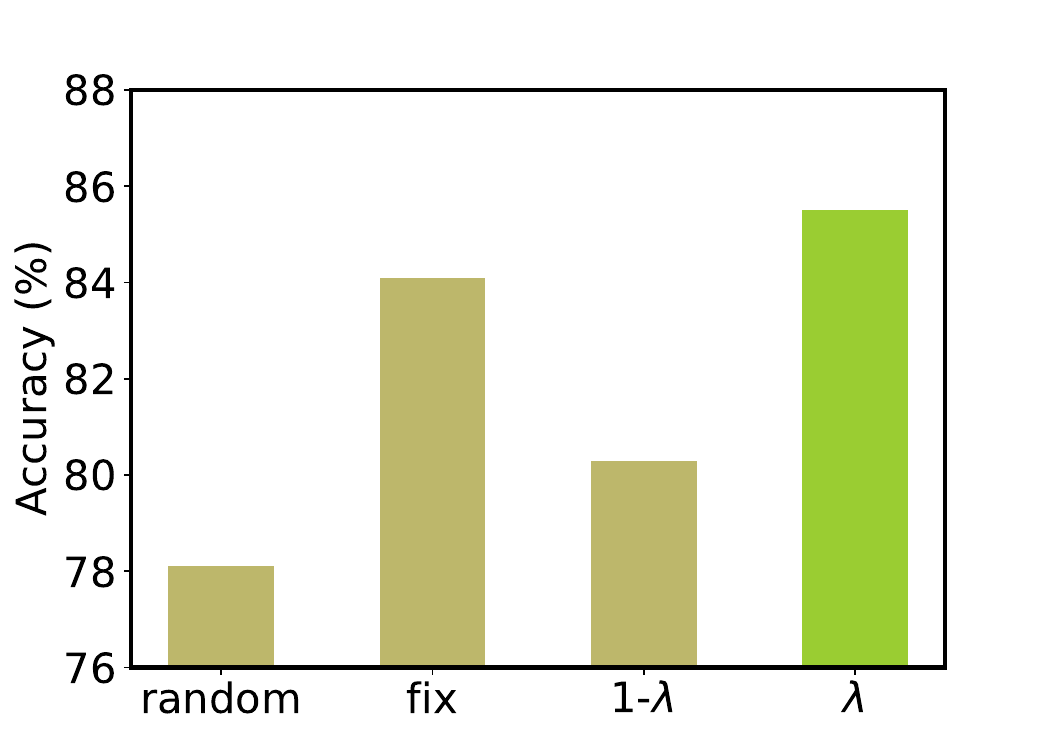}}
  \centerline{(a) Different strategies for $\lambda$}
\end{minipage}
\hfill
\begin{minipage}[b]{0.40\linewidth}
  \centering
\centerline{\includegraphics[width=3.8cm]{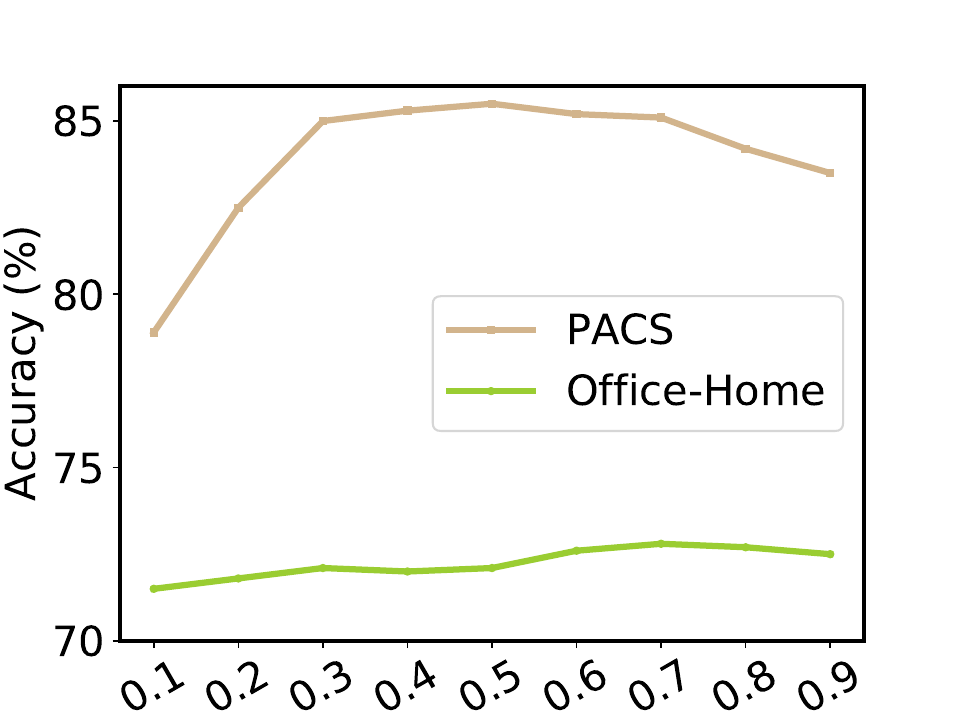}}
  \centerline{(b) Initial value $\lambda_0$}
\end{minipage}
\caption{\textbf{Parameter $\lambda$ Analysis on PACS.} (a) The strategy in Eq.~(\ref{eq:lambda}) achieves best performance. (b) Different tasks require different initial values. }
\label{fig:lambda}
\end{figure}

\begin{figure}[t!]
\begin{minipage}[b]{1.0\linewidth}
  \centering
\centerline{\includegraphics[width=8.5cm]{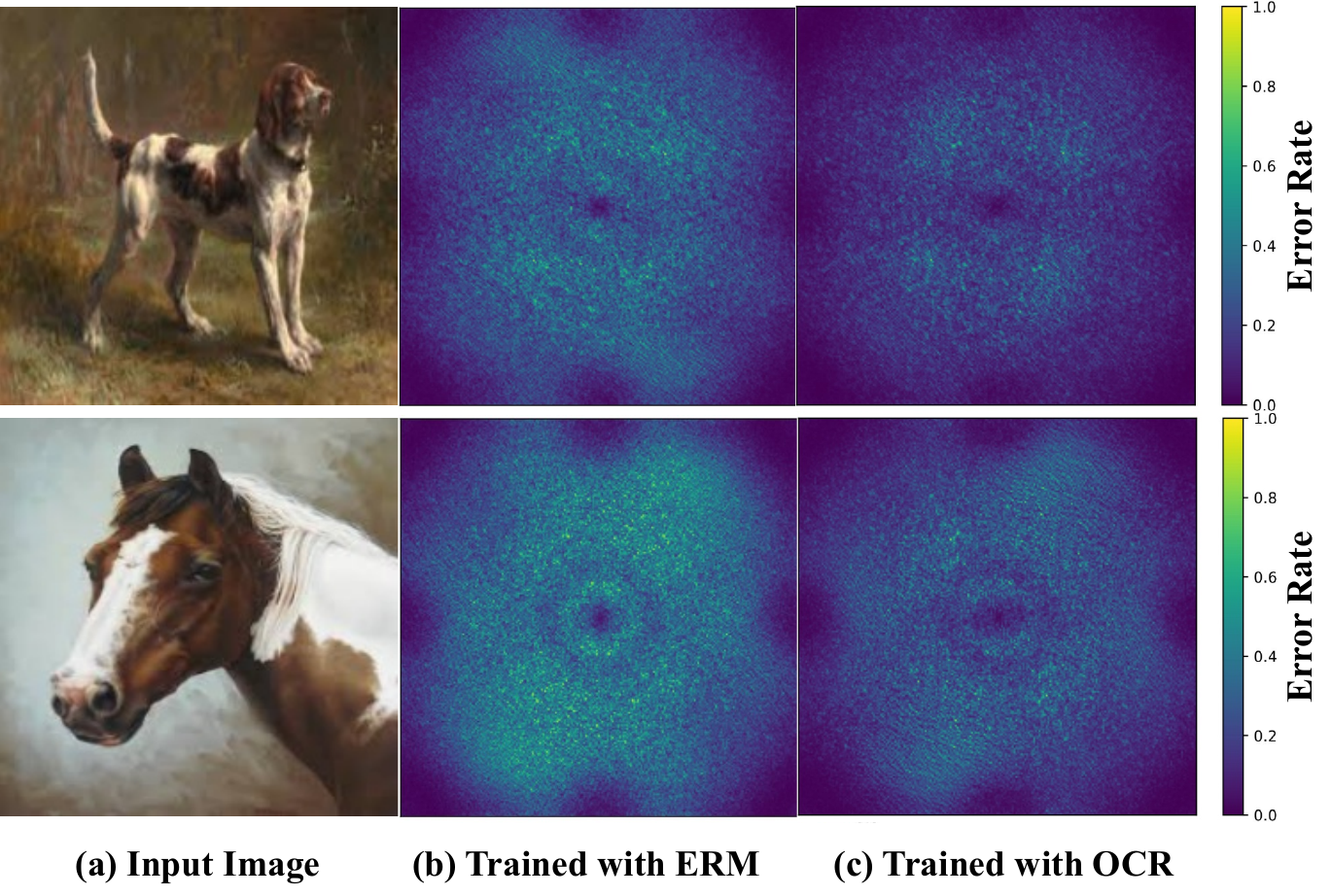}}
\end{minipage}
\caption{\textbf{Fourier Perspective.} Model sensitivity to additive noise aligned with different Fourier basis vectors on PACS (Art). The pixels closer to the center in the heat map represent the impact of low frequency noise, while the pixels outward represents the impact of high frequency noise. The model trained with OCR is more robust compared with the model learned by ERM.}
\label{fig:freq}
\vspace{-10pt}
\end{figure}

\begin{table}[t]
\footnotesize
\centering
\caption{
\textbf{Analysis on Other Layers.} 
Accuracies (\%) on Office-Home, where ``Input" means we apply OCR on pixel-level and ``BT*" is the bottleneck block of ResNet-50. Generally, the deeper the layer, the more effective OCR will be.}
\vspace{5pt}
\label{tab:diff_layers}
\resizebox{\columnwidth}{!}{%
\begin{tabular}{ccccccc}
\toprule
\multicolumn{7}{c}{\textbf{ResNet-50 layers}}\\
	\cmidrule(lr){1-7}
Input & Conv1 & BT1  & BT2  & BT3  & BT4  & FC  \\\midrule
68.5  & 69.0  & 70.4 & 70.6 & 71.2 & 72.2 & 72.8 \\ \bottomrule
\end{tabular}%
}
\end{table}

\textbf{Domain Generalization Classification.} 
In Table~\ref{tab:dgc}, OCR outperforms the vanilla ResNet-18 with a large margin. Note that Mixup and Manifold Mixup do not improve the vanilla ResNet-18. The reason why Mixup is ineffective here is because Mixup mainly encourages the model to be robust to the combination of the existing patterns, but does not enhance the ability to handle the unseen styles. MixStyle regularizes the model to be robust to the unseen styles, however, it does not explicitly minimize the domain-specific information in the representation, leading to its worse performance than OCR. We also test our method on PACS based on ResNet-50 and SWAD~\cite{cha2021swad}. OCR improves SWAD from 87.8\% to 88.5\%. OCR achieves consistent performance advantages on both ResNet-18 and ResNet-50.
 
 \textbf{Domain Generalization Semantic Segmentation.}
In Table~\ref{tab:segment}, the Baseline is DeepLabV3+~\cite{chen2018encoder}. OCR improves the Baseline by 6.5\%. For IBN-Net~\cite{pan2018two}, the improvement is 2.7\%, which is also impressive. For RobustNet~\cite{choi2021robustnet}, we observe that OCR has a small improvement of 0.6\%, this may be because RobustNet also enhances the generalization of the model by eliminating domain specific information. OCR, however, is different from RobustNet since RobustNet disentangles the domain-specific and domain-invariant part in the feature covariance while OCR does this based on the assumption of linear combination.

\textbf{Cross-domain Object Detection.} In Table~\ref{tab:od}, we report results of OCR and the compared methods on the object detection task of City~\cite{cordts2016cityscapes}$\to$ FoggyCity~\cite{sakaridis2018semantic}. OCR achieves improvements of 1.6\% and 1.4\% compared to SDAT~\cite{rangwani2022closer} and SUDA~\cite{zhang2022spectral}, respectively. Therefore, OCR is effective on object detection tasks.

\subsection{Ablations}
\textbf{Parameter $\lambda$ Analysis.}
In Fig.~\ref{fig:lambda}, we provide an analysis for parameter $\lambda$. In Fig.~\ref{fig:lambda} (a), we illustrate the impact of different choices for $\lambda$ on PACS, where ``random'' indicates we choose a random value from range (0,1) as $\lambda$ during each iteration, ``fix'' means we fix $\lambda$ as 0.5, ``$\lambda$'' represents the strategy in Eq.~(\ref{eq:lambda}), while ``$1-\lambda$'' is the opposite strategy of ``$\lambda$''. As can be seen from Fig.~\ref{fig:lambda} (a), using the strategy in Eq.~(\ref{eq:lambda}) helps to train an optimal model. This result is in line with our hypothesis, \ie, at the beginning of training, there is a small domain-specific ratio in the representation, so a small $\lambda$ is required, as OCR continuously minimizes the domain-specific information, the domain-invariant part gradually increases, so a larger $\lambda$ is required. 
In addition, we test the performance of OCR with the simple formulation, i.e., $z_n {=} z_a {-} z _o$, on PACS and achieve an accuracy of 84.0\%, which is close to that of the fixed proportion setting in Fig.~\ref{fig:lambda} (a), but lower than our formulation which obtains 85.5\%.
In Fig.~\ref{fig:lambda} (b), we report the impact of different initial values $\lambda_0$ on performance. From Fig.~\ref{fig:lambda} (b), we observe that the results of Office-Home do not fluctuate much with different intial values, its best $\lambda_0$ is around 0.7. The results of PACS, are more sensitive to different initial values, its best $\lambda_0$ is around 0.5.
Therefore, different tasks need different initial values.

\textbf{Analysis on Other Layers.} By default we apply OCR to representations of the penultimate layer of the model. OCR can be applied to the representations of other layers as well. We show results with a ResNet-50 in Table~\ref{tab:diff_layers}, we observe that: (1) In the image level, OCR cannot achieve ideal results, which may be because some attributes of the sample, \eg, lighting and shooting angle, cannot be separated in the image level; (2) In general, the deeper the layer, the more effective OCR will be. Prior works~\cite{long2018transferable,yosinski2014transferable} have found that representations extracted from the shallow layers are more generalized, while the representations extracted from the deep layers show strong task relevance. Therefore, shallow representations are not suitable for applying OCR, while deep representations can eliminate domain-specific information through OCR.

\textbf{Fourier Perspective.} Following \cite{yin2019fourier}, we investigate the sensitivity of our models to high and low frequency corruptions via a perturbation analysis in the Fourier domain. We plot the Fourier heat map in Fig.~\ref{fig:freq}. The pixels closer to the center in the heat map represent the impact of low frequency noise, while the pixels outward represent the impact of high frequency noise. We observe that the model trained with OCR is more robust compared with the model learned by ERM, especially in the high frequency domain. High frequency information is often introduced by styles that vary significant across domains. Therefore, OCR can effectively eliminates the domain-specific style information.

\textbf{Robustness to Adversarial Attack.}
In Table~\ref{tab:attack} we report the adversarial robustness of our method against various white-box attacks, including FGSM~\cite{good2015explain}, BIM~\cite{kurakin2017adversarial} and PGD~\cite{madry2018towards}. We impose the adversarial attacks through the Adversarial Robust Tool box\footnote{\url{https://github.com/advboxes/AdvBox}}. For fair comparison, we set the iteration number as 10, adversarial strength as 0.01 and step size as 0.01, all other parameters remain at their default values. 
Compared with ERM and prediction-based consistency regularization, OCR achieves the best robustness to all the three adversarial attacks. 
Especially for the iterative-based methods with more powerful attacks, OCR achieves accuracies of 61.6\% and 61.7\% against PGD and BIM, which is remarkably higher than ERM and prediction-based consistency regularization. The superior robustness of OCR against the adversarial attack derives from explicitly eliminating the negative effects of the domain-specific attributes which causes the domain shifts.

\textbf{Effect of Order-preserving Property.} In Table~\ref{tab:top5_acc}, we report the top 1 to top 5 accuracies on Office-Home. Compared with prediction-based method, OCR has more significant advantages in top 3 and top 5 accuracies, which proves that the order-preserving property in consistency regularization guarantees that even though the maximum probability category does not hit the ground-truth label, it is very likely that the label appears in the top 3 or top 5 categories.

\textbf{Effect of Data Augmentations.} Following the setting in Table~\ref{tab:attack}, we remove ColorJitter, RandomGrayscale and GaussianBlur, respectively. The results are reported in Table~\ref{tab:effect_aug}. We observe that the combination of three augmentations can achieve the best performance.
According to the practice in self-supervised learning~\cite{chen2020simple}, not all the combinations help improve the generalization of the model. Exploring the best combination would be a promising future work.

\begin{table}[t]
\Huge
\centering
\caption{\textbf{Robustness to Adversarial Attack.} Accuracy (\%) on PACS after different adversarial attacks. The results are all based on the leave-one-domain-out protocol \cite{zhou2020domain}. Our method is effective to enhance the model robustness to the adversarial attacks.}
\label{tab:attack}
\vspace{5pt}
\resizebox{1\columnwidth}{!}{%
\begin{tabular}{lcccccr}
\toprule
& \multicolumn{5}{c}{\textbf{PACS}}\\
	\cmidrule(lr){2-6}
& Art & Cartoon & Photo & Sketch & Mean & Boost \\\midrule
\rowcolor{mygray}
\textbf{No attack} & & & & & &\\
ERM Baseline         & 78.3          & 76.0    & 95.0  & 72.7   & 80.5 & \\
\quad w/ P-Cons. Reg. & 79.2          & 80.2    & 95.9  & 79.3   & 83.7 & 3.2 $\uparrow$ \\
\quad w/ OCR & {\bf 85.4}    & {\bf 81.1}    & {\bf 96.2}  & {\bf 81.2}   & {\bf 86.0} & {\bf 5.5 $\uparrow$}\\
\rowcolor{mygray}
\textbf{FGSM attack}~\cite{good2015explain} & & & & & &\\
ERM Baseline         & 19.4          & 55.8    & 51.5  & 48.7   & 43.9 &       \\
\quad w/ P-Cons. Reg. & 32.0          & 63.8    & 66.2  & 66.6   & 57.2 & 13.3 $\uparrow$     \\
\quad w/ OCR & {\bf 45.5}          & {\bf 69.5}    & {\bf 73.3}  & {\bf 73.2}   & {\bf 65.4} & {\bf 21.5 $\uparrow$} \\
\rowcolor{mygray}
\textbf{BIM attack}~\cite{kurakin2017adversarial}  & & & & & &\\
ERM Baseline         & 16.6          & 55.0    & 40.5  & 41.6   & 38.4 &       \\
\quad w/ P-Cons. Reg. & 26.5          & 63.3    & 59.9  & 60.6   & 52.6 & 14.2 $\uparrow$      \\
\quad w/ OCR & {\bf 38.7}          & {\bf 69.1}    & {\bf 68.4}  & {\bf 70.4}   & {\bf 61.7} & {\bf 23.3 $\uparrow$}  \\
\rowcolor{mygray}
\textbf{PGD attack}~\cite{madry2018towards}  & & & & & &\\
ERM Baseline         & 16.8       & 54.9    & 40.4  & 41.5   & 38.4 &       \\
\quad w/ P-Cons. Reg. & 26.1       & 63.4    & 60.1  & 60.8   & 52.6 & 14.2 $\uparrow$     \\
\quad w/ OCR   & {\bf 38.5} & {\bf 69.0}    & {\bf 68.2}  & {\bf 70.6}   & {\bf 61.6} & {\bf 23.2 $\uparrow$}  \\\bottomrule
\end{tabular}%
}
\vspace{-5pt}
\end{table}

\begin{table}[t]
\Huge
\centering
\caption{\bf Top 1 to top 5 accuracies (\%) on Office-Home. Baseline method is SHOT.}
\vspace{5pt}
\label{tab:top5_acc}
\resizebox{.48\textwidth}{!}{%
\begin{tabular}{cccc}
\toprule
Acc.(\%) & Baseline & Baseline+P-Cons. Reg. & Baseline+OCR \\ \midrule
Top 1    & 71.8     & 72.0                  & 72.8         \\
Top 3    & 85.5     & 86.1                  & 87.6         \\
Top 5    & 89.4     & 90.2                  & 92.2         \\ \bottomrule
\end{tabular}%
}
\end{table}

\begin{table}[t]
\Huge
\centering
\caption{\textbf{Effect of data augmentations.}}
\vspace{5pt}
\label{tab:effect_aug}
\resizebox{.48\textwidth}{!}{%
\begin{tabular}{lcccc}
\toprule
&  & \multicolumn{3}{c}{Removing one augmentation} \\
\cmidrule(lr){3-5}
Tasks                        & All  & ColorJitter & RandomGray. & GaussianBlur \\ \midrule
PACS (Ours)        & 85.5 & 83.4 & 84.6 & 83.9             \\ 
PACS (P-Cons. Reg.)  & 83.7 & 82.0 & 83.1 & 82.7             \\ 
\midrule
+PGD (Ours) & 61.6 & 52.9 & 55.3 & 47.1             \\ 
+PGD (P-Cons. Reg.) & 52.6 & 45.2 & 47.6 & 41.5        \\ 
\bottomrule
\end{tabular}%
}
\vspace{-10pt}
\end{table}

\section{Conclusion and Future work}
In this paper, we propose Order-preserving Consistency Regularization (OCR) to enhance model robustness to domain-specific attributes for cross-domain tasks. 
We first separate the residual component from the augmented representation.
Then, we maximize the entropy of the residual component to enlarge the uncertainty of its prediction.
As a result, the residual component contains little information about the task of interest, \ie, the model is less sensitive to the domain-specific attributes.
Throughout the experiments, we have shown that OCR enhances the generalization of the model and provides better robustness to adversarial attacks. OCR is easy to implement and can be applied to any cross-domain task to improve the performance.
Like any data-augmentation based method, our proposal fails when the augmentations are completely independent of the domain gaps. Therefore, exploring the most related data augmentations for specific cross-domain tasks would be a suitable future work.

\section*{Acknowledgment}
This work was supported in part by the National Natural Science Foundation of China under Grant 62250061, 62176042, 62276054, and in part by the Sichuan Science and Technology Program under Grant 2023YFG0156, and in part by CAAI-Huawei MindSpore Open Fund.

{\small
\bibliographystyle{ieee_fullname}
\bibliography{myreference}

\begin{thebibliography}{10}\itemsep=-1pt

\bibitem{achille2018emergence}
Alessandro Achille and Stefano Soatto.
\newblock Emergence of invariance and disentanglement in deep representations.
\newblock {\em Journal of Machine Learning Research}, 19(1):1947--1980, 2018.

\bibitem{bachman2014learning}
Philip Bachman, Ouais Alsharif, and Doina Precup.
\newblock Learning with pseudo-ensembles.
\newblock In {\em Advances in Neural Information Processing Systems},
  volume~27, 2014.

\bibitem{balaji2018metareg}
Yogesh Balaji, Swami Sankaranarayanan, and Rama Chellappa.
\newblock Metareg: Towards domain generalization using meta-regularization.
\newblock In {\em Advances in Neural Information Processing Systems},
  volume~31, 2018.

\bibitem{berthelot2019remixmatch}
David Berthelot, Nicholas Carlini, Ekin~D Cubuk, Alex Kurakin, Kihyuk Sohn, Han
  Zhang, and Colin Raffel.
\newblock Remixmatch: Semi-supervised learning with distribution matching and
  augmentation anchoring.
\newblock In {\em International Conference on Learning Representations}, 2019.

\bibitem{berthelot2019mixmatch}
David Berthelot, Nicholas Carlini, Ian Goodfellow, Nicolas Papernot, Avital
  Oliver, and Colin~A Raffel.
\newblock Mixmatch: A holistic approach to semi-supervised learning.
\newblock In {\em Advances in Neural Information Processing Systems},
  volume~32, 2019.

\bibitem{carlucci2019domain}
Fabio~M Carlucci, Antonio D'Innocente, Silvia Bucci, Barbara Caputo, and
  Tatiana Tommasi.
\newblock Domain generalization by solving jigsaw puzzles.
\newblock In {\em IEEE/CVF Conference on Computer Vision and Pattern
  Recognition}, pages 2229--2238, 2019.

\bibitem{cha2021swad}
Junbum Cha, Sanghyuk Chun, Kyungjae Lee, Han-Cheol Cho, Seunghyun Park, Yunsung
  Lee, and Sungrae Park.
\newblock Swad: Domain generalization by seeking flat minima.
\newblock In {\em Advances in Neural Information Processing Systems},
  volume~34, pages 22405--22418, 2021.

\bibitem{chen2022contrastive}
Dian Chen, Dequan Wang, Trevor Darrell, and Sayna Ebrahimi.
\newblock Contrastive test-time adaptation.
\newblock In {\em IEEE/CVF Conference on Computer Vision and Pattern
  Recognition}, pages 295--305, 2022.

\bibitem{chen2018encoder}
Liang-Chieh Chen, Yukun Zhu, George Papandreou, Florian Schroff, and Hartwig
  Adam.
\newblock Encoder-decoder with atrous separable convolution for semantic image
  segmentation.
\newblock In {\em European Conference on Computer Vision}, pages 801--818,
  2018.

\bibitem{chen2020simple}
Ting Chen, Simon Kornblith, Mohammad Norouzi, and Geoffrey Hinton.
\newblock A simple framework for contrastive learning of visual
  representations.
\newblock In {\em International Conference on Machine Learning}, pages
  1597--1607. PMLR, 2020.

\bibitem{chen2021exploring}
Xinlei Chen and Kaiming He.
\newblock Exploring simple siamese representation learning.
\newblock In {\em IEEE/CVF Conference on Computer Vision and Pattern
  Recognition}, pages 15750--15758, 2021.

\bibitem{choi2021robustnet}
Sungha Choi, Sanghun Jung, Huiwon Yun, Joanne~T Kim, Seungryong Kim, and Jaegul
  Choo.
\newblock Robustnet: Improving domain generalization in urban-scene
  segmentation via instance selective whitening.
\newblock In {\em IEEE/CVF Conference on Computer Vision and Pattern
  Recognition}, pages 11580--11590, 2021.

\bibitem{cordts2016cityscapes}
Marius Cordts, Mohamed Omran, Sebastian Ramos, Timo Rehfeld, Markus Enzweiler,
  Rodrigo Benenson, Uwe Franke, Stefan Roth, and Bernt Schiele.
\newblock The cityscapes dataset for semantic urban scene understanding.
\newblock In {\em IEEE/CVF Conference on Computer Vision and Pattern
  Recognition}, pages 3213--3223, 2016.

\bibitem{devries2017improved}
Terrance DeVries and Graham~W Taylor.
\newblock Improved regularization of convolutional neural networks with cutout.
\newblock {\em arXiv preprint arXiv:1708.04552}, 2017.

\bibitem{ganin2015unsupervised}
Yaroslav Ganin and Victor Lempitsky.
\newblock Unsupervised domain adaptation by backpropagation.
\newblock In {\em International Conference on Machine Learning}, pages
  1180--1189. PMLR, 2015.

\bibitem{ganin2016domain}
Yaroslav Ganin, Evgeniya Ustinova, Hana Ajakan, Pascal Germain, Hugo
  Larochelle, Fran{\c{c}}ois Laviolette, Mario Marchand, and Victor Lempitsky.
\newblock Domain-adversarial training of neural networks.
\newblock {\em Journal of Machine Learning Research}, 17(1):2096--2030, 2016.

\bibitem{ghiasi2018dropblock}
Golnaz Ghiasi, Tsung-Yi Lin, and Quoc~V Le.
\newblock Dropblock: A regularization method for convolutional networks.
\newblock In {\em Advances in Neural Information Processing Systems},
  volume~31, 2018.

\bibitem{ghifary2015domain}
Muhammad Ghifary, W~Bastiaan Kleijn, Mengjie Zhang, and David Balduzzi.
\newblock Domain generalization for object recognition with multi-task
  autoencoders.
\newblock In {\em IEEE International Conference on Computer Vision}, pages
  2551--2559, 2015.

\bibitem{gong2019dlow}
Rui Gong, Wen Li, Yuhua Chen, and Luc~Van Gool.
\newblock Dlow: Domain flow for adaptation and generalization.
\newblock In {\em IEEE/CVF Conference on Computer Vision and Pattern
  Recognition}, pages 2477--2486, 2019.

\bibitem{good2015explain}
Ian Goodfellow, Jonathon Shlens, and Christian Szegedy.
\newblock Explaining and harnessing adversarial examples.
\newblock In {\em International Conference on Learning Representations}, 2015.

\bibitem{grill2020bootstrap}
Jean-Bastien Grill, Florian Strub, Florent Altch{\'e}, Corentin Tallec, Pierre
  Richemond, Elena Buchatskaya, Carl Doersch, Bernardo Avila~Pires, Zhaohan
  Guo, and Mohammad Gheshlaghi~Azar.
\newblock Bootstrap your own latent-a new approach to self-supervised learning.
\newblock In {\em Advances in Neural Information Processing Systems},
  volume~33, pages 21271--21284, 2020.

\bibitem{guan2021domain}
Hao Guan and Mingxia Liu.
\newblock Domain adaptation for medical image analysis: a survey.
\newblock {\em IEEE Transactions on Biomedical Engineering}, 69(3):1173--1185,
  2021.

\bibitem{he2016deep}
Kaiming He, Xiangyu Zhang, Shaoqing Ren, and Jian Sun.
\newblock Deep residual learning for image recognition.
\newblock In {\em IEEE/CVF Conference on Computer Vision and Pattern
  Recognition}, pages 770--778, 2016.

\bibitem{hendrycks2018benchmarking}
Dan Hendrycks and Thomas Dietterich.
\newblock Benchmarking neural network robustness to common corruptions and
  perturbations.
\newblock In {\em International Conference on Learning Representations}, 2019.

\bibitem{hendrycks2020augmix}
Dan Hendrycks, Norman Mu, Ekin~Dogus Cubuk, Barret Zoph, Justin Gilmer, and
  Balaji Lakshminarayanan.
\newblock Augmix: A simple method to improve robustness and uncertainty under
  data shift.
\newblock In {\em International Conference on Learning Representations}, 2020.

\bibitem{huang2021model}
Jiaxing Huang, Dayan Guan, Aoran Xiao, and Shijian Lu.
\newblock Model adaptation: Historical contrastive learning for unsupervised
  domain adaptation without source data.
\newblock In {\em Advances in Neural Information Processing Systems},
  volume~34, 2021.

\bibitem{islam2021dynamic}
Ashraful Islam, Chun-Fu~Richard Chen, Rameswar Panda, Leonid Karlinsky, Rogerio
  Feris, and Richard~J Radke.
\newblock Dynamic distillation network for cross-domain few-shot recognition
  with unlabeled data.
\newblock In {\em Advances in Neural Information Processing Systems},
  volume~34, pages 3584--3595, 2021.

\bibitem{jing2022variational}
Mengmeng Jing, Xiantong Zhen, Jingjing Li, and Cees G.~M. Snoek.
\newblock Variational model perturbation for source-free domain adaptation.
\newblock In {\em Advances in Neural Information Processing Systems}, 2022.

\bibitem{kim2021selfreg}
Daehee Kim, Youngjun Yoo, Seunghyun Park, Jinkyu Kim, and Jaekoo Lee.
\newblock Selfreg: Self-supervised contrastive regularization for domain
  generalization.
\newblock In {\em IEEE/CVF International Conference on Computer Vision}, pages
  9619--9628, 2021.

\bibitem{krizhevsky2009learning}
Alex Krizhevsky and Geoffrey Hinton.
\newblock Learning multiple layers of features from tiny images.
\newblock {\em Technical report}, 2009.

\bibitem{kurakin2017adversarial}
Alexey Kurakin, Ian~J. Goodfellow, and Samy Bengio.
\newblock Adversarial machine learning at scale.
\newblock In {\em International Conference on Learning Representations}, 2017.

\bibitem{laine2017temporal}
Samuli Laine and Timo Aila.
\newblock Temporal ensembling for semi-supervised learning.
\newblock In {\em International Conference on Learning Representations}, 2017.

\bibitem{li2022invariant}
Bo Li, Yifei Shen, Yezhen Wang, Wenzhen Zhu, Dongsheng Li, Kurt Keutzer, and
  Han Zhao.
\newblock Invariant information bottleneck for domain generalization.
\newblock In {\em AAAI Conference on Artificial Intelligence}, volume~36, pages
  7399--7407, 2022.

\bibitem{li2017deeper}
Da Li, Yongxin Yang, Yi-Zhe Song, and Timothy~M Hospedales.
\newblock Deeper, broader and artier domain generalization.
\newblock In {\em IEEE International Conference on Computer Vision}, pages
  5542--5550, 2017.

\bibitem{li2018domain}
Haoliang Li, Sinno~Jialin Pan, Shiqi Wang, and Alex~C Kot.
\newblock Domain generalization with adversarial feature learning.
\newblock In {\em IEEE/CVF Conference on Computer Vision and Pattern
  Recognition}, pages 5400--5409, 2018.

\bibitem{li2021simple}
Pan Li, Da Li, Wei Li, Shaogang Gong, Yanwei Fu, and Timothy~M Hospedales.
\newblock A simple feature augmentation for domain generalization.
\newblock In {\em IEEE/CVF International Conference on Computer Vision}, pages
  8886--8895, 2021.

\bibitem{li2017revisiting}
Yanghao Li, Naiyan Wang, Jianping Shi, Jiaying Liu, and Xiaodi Hou.
\newblock Revisiting batch normalization for practical domain adaptation.
\newblock In {\em International Conference on Learning Representations
  Workshop}, 2017.

\bibitem{liang2020we}
Jian Liang, Dapeng Hu, and Jiashi Feng.
\newblock Do we really need to access the source data? source hypothesis
  transfer for unsupervised domain adaptation.
\newblock In {\em International Conference on Machine Learning}, pages
  6028--6039. PMLR, 2020.

\bibitem{liang2021source}
Jian Liang, Dapeng Hu, Yunbo Wang, Ran He, and Jiashi Feng.
\newblock Source data-absent unsupervised domain adaptation through hypothesis
  transfer and labeling transfer.
\newblock {\em IEEE Transactions on Pattern Analysis and Machine Intelligence},
  44(11):8602--8617, 2021.

\bibitem{long2018transferable}
Mingsheng Long, Yue Cao, Zhangjie Cao, Jianmin Wang, and Michael~I Jordan.
\newblock Transferable representation learning with deep adaptation networks.
\newblock {\em IEEE Transactions on Pattern Analysis and Machine Intelligence},
  41(12):3071--3085, 2018.

\bibitem{long2018conditional}
Mingsheng Long, Zhangjie Cao, Jianmin Wang, and Michael~I Jordan.
\newblock Conditional adversarial domain adaptation.
\newblock In {\em Advances in Neural Information Processing Systems}, pages
  1640--1650, 2018.

\bibitem{lv2022causality}
Fangrui Lv, Jian Liang, Shuang Li, Bin Zang, Chi~Harold Liu, Ziteng Wang, and
  Di Liu.
\newblock Causality inspired representation learning for domain generalization.
\newblock In {\em IEEE/CVF Conference on Computer Vision and Pattern
  Recognition}, pages 8046--8056, 2022.

\bibitem{madry2018towards}
Aleksander Madry, Aleksandar Makelov, Ludwig Schmidt, Dimitris Tsipras, and
  Adrian Vladu.
\newblock Towards deep learning models resistant to adversarial attacks.
\newblock In {\em International Conference on Learning Representations}, 2018.

\bibitem{miyato2018virtual}
Takeru Miyato, Shin-ichi Maeda, Masanori Koyama, and Shin Ishii.
\newblock Virtual adversarial training: a regularization method for supervised
  and semi-supervised learning.
\newblock {\em IEEE Transactions on Pattern Analysis and Machine Intelligence},
  41(8):1979--1993, 2018.

\bibitem{motiian2017unified}
Saeid Motiian, Marco Piccirilli, Donald~A Adjeroh, and Gianfranco Doretto.
\newblock Unified deep supervised domain adaptation and generalization.
\newblock In {\em IEEE International Conference on Computer Vision}, pages
  5715--5725, 2017.

\bibitem{muandet2013domain}
Krikamol Muandet, David Balduzzi, and Bernhard Sch{\"o}lkopf.
\newblock Domain generalization via invariant feature representation.
\newblock In {\em International Conference on Machine Learning}, pages 10--18.
  PMLR, 2013.

\bibitem{neuhold2017mapillary}
Gerhard Neuhold, Tobias Ollmann, Samuel Rota~Bulo, and Peter Kontschieder.
\newblock The mapillary vistas dataset for semantic understanding of street
  scenes.
\newblock In {\em IEEE International Conference on Computer Vision}, pages
  4990--4999, 2017.

\bibitem{niu2015multi}
Li Niu, Wen Li, and Dong Xu.
\newblock Multi-view domain generalization for visual recognition.
\newblock In {\em IEEE International Conference on Computer Vision}, pages
  4193--4201, 2015.

\bibitem{niu2015visual}
Li Niu, Wen Li, and Dong Xu.
\newblock Visual recognition by learning from web data: A weakly supervised
  domain generalization approach.
\newblock In {\em IEEE/CVF Conference on Computer Vision and Pattern
  Recognition}, pages 2774--2783, 2015.

\bibitem{pan2009survey}
Sinno~Jialin Pan and Qiang Yang.
\newblock A survey on transfer learning.
\newblock {\em IEEE Transactions on knowledge and data engineering},
  22(10):1345--1359, 2009.

\bibitem{pan2018two}
Xingang Pan, Ping Luo, Jianping Shi, and Xiaoou Tang.
\newblock Two at once: Enhancing learning and generalization capacities via
  ibn-net.
\newblock In {\em European Conference on Computer Vision}, pages 464--479,
  2018.

\bibitem{paszke2019pytorch}
Adam Paszke, Sam Gross, Francisco Massa, Adam Lerer, James Bradbury, Gregory
  Chanan, Trevor Killeen, Zeming Lin, Natalia Gimelshein, Luca Antiga, et~al.
\newblock Pytorch: An imperative style, high-performance deep learning library.
\newblock In {\em Advances in Neural Information Processing Systems},
  volume~32, 2019.

\bibitem{rangwani2022closer}
Harsh Rangwani, Sumukh~K Aithal, Mayank Mishra, Arihant Jain, and
  Venkatesh~Babu Radhakrishnan.
\newblock A closer look at smoothness in domain adversarial training.
\newblock In {\em International Conference on Machine Learning}, pages
  18378--18399. PMLR, 2022.

\bibitem{richter2016playing}
Stephan~R Richter, Vibhav Vineet, Stefan Roth, and Vladlen Koltun.
\newblock Playing for data: Ground truth from computer games.
\newblock In {\em European Conference on Computer Vision}, pages 102--118.
  Springer, 2016.

\bibitem{ros2016synthia}
German Ros, Laura Sellart, Joanna Materzynska, David Vazquez, and Antonio~M
  Lopez.
\newblock The synthia dataset: A large collection of synthetic images for
  semantic segmentation of urban scenes.
\newblock In {\em IEEE/CVF Conference on Computer Vision and Pattern
  Recognition}, pages 3234--3243, 2016.

\bibitem{saito2018maximum}
Kuniaki Saito, Kohei Watanabe, Yoshitaka Ushiku, and Tatsuya Harada.
\newblock Maximum classifier discrepancy for unsupervised domain adaptation.
\newblock In {\em IEEE/CVF Conference on Computer Vision and Pattern
  Recognition}, pages 3723--3732, 2018.

\bibitem{sajjadi2016regularization}
Mehdi Sajjadi, Mehran Javanmardi, and Tolga Tasdizen.
\newblock Regularization with stochastic transformations and perturbations for
  deep semi-supervised learning.
\newblock In {\em Advances in Neural Information Processing Systems},
  volume~29, 2016.

\bibitem{sakaridis2018semantic}
Christos Sakaridis, Dengxin Dai, and Luc Van~Gool.
\newblock Semantic foggy scene understanding with synthetic data.
\newblock {\em International Journal of Computer Vision}, 126:973--992, 2018.

\bibitem{sohn2020fixmatch}
Kihyuk Sohn, David Berthelot, Nicholas Carlini, Zizhao Zhang, Han Zhang,
  Colin~A Raffel, Ekin~Dogus Cubuk, Alexey Kurakin, and Chun-Liang Li.
\newblock Fixmatch: Simplifying semi-supervised learning with consistency and
  confidence.
\newblock In {\em Advances in Neural Information Processing Systems},
  volume~33, pages 596--608, 2020.

\bibitem{sugiyama2007covariate}
Masashi Sugiyama, Matthias Krauledat, and Klaus-Robert M{\~A}{\v{z}}ller.
\newblock Covariate shift adaptation by importance weighted cross validation.
\newblock {\em Journal of Machine Learning Research}, 8(May):985--1005, 2007.

\bibitem{tarvainen2017mean}
Antti Tarvainen and Harri Valpola.
\newblock Mean teachers are better role models: Weight-averaged consistency
  targets improve semi-supervised deep learning results.
\newblock In {\em Advances in Neural Information Processing Systems},
  volume~30, 2017.

\bibitem{tishby1999information}
Naftali Tishby, Fernando~C. Pereira, and William Bialek.
\newblock The information bottleneck method.
\newblock In {\em Annual Allerton Conference on Communications, Control and
  Computing}, pages 368--377, 1999.

\bibitem{tzeng2017adversarial}
Eric Tzeng, Judy Hoffman, Kate Saenko, and Trevor Darrell.
\newblock Adversarial discriminative domain adaptation.
\newblock In {\em IEEE/CVF Conference on Computer Vision and Pattern
  Recognition}, pages 7167--7176, 2017.

\bibitem{Vapnik1998}
Vladimir~N. Vapnik.
\newblock {\em Statistical Learning Theory}.
\newblock Wiley-Interscience, 1998.

\bibitem{venkateswara2017deep}
Hemanth Venkateswara, Jose Eusebio, Shayok Chakraborty, and Sethuraman
  Panchanathan.
\newblock Deep hashing network for unsupervised domain adaptation.
\newblock In {\em IEEE/CVF Conference on Computer Vision and Pattern
  Recognition}, pages 5018--5027, 2017.

\bibitem{verma2019manifold}
Vikas Verma, Alex Lamb, Christopher Beckham, Amir Najafi, Ioannis Mitliagkas,
  David Lopez-Paz, and Yoshua Bengio.
\newblock Manifold mixup: Better representations by interpolating hidden
  states.
\newblock In {\em International Conference on Machine Learning}, pages
  6438--6447. PMLR, 2019.

\bibitem{wang2020tent}
Dequan Wang, Evan Shelhamer, Shaoteng Liu, Bruno Olshausen, and Trevor Darrell.
\newblock Tent: Fully test-time adaptation by entropy minimization.
\newblock In {\em International Conference on Learning Representations}, 2020.

\bibitem{wang2022continual}
Qin Wang, Olga Fink, Luc Van~Gool, and Dengxin Dai.
\newblock Continual test-time domain adaptation.
\newblock In {\em IEEE/CVF Conference on Computer Vision and Pattern
  Recognition}, 2022.

\bibitem{wiles2021fine}
Olivia Wiles, Sven Gowal, Florian Stimberg, Sylvestre-Alvise Rebuffi, Ira
  Ktena, Krishnamurthy~Dj Dvijotham, and Ali~Taylan Cemgil.
\newblock A fine-grained analysis on distribution shift.
\newblock In {\em International Conference on Learning Representations}, 2021.

\bibitem{xiao2021learning}
Zehao Xiao, Xiantong Zhen, Ling Shao, and Cees~GM Snoek.
\newblock Learning to generalize across domains on single test samples.
\newblock In {\em International Conference on Learning Representations}, 2021.

\bibitem{xie2020unsupervised}
Qizhe Xie, Zihang Dai, Eduard Hovy, Thang Luong, and Quoc Le.
\newblock Unsupervised data augmentation for consistency training.
\newblock In {\em Advances in Neural Information Processing Systems},
  volume~33, pages 6256--6268, 2020.

\bibitem{yang2021exploiting}
Shiqi Yang, Yaxing Wang, Joost van~de Weijer, Luis Herranz, and Shangling Jui.
\newblock Exploiting the intrinsic neighborhood structure for source-free
  domain adaptation.
\newblock In {\em Advances in Neural Information Processing Systems},
  volume~34, 2021.

\bibitem{yin2019fourier}
Dong Yin, Raphael Gontijo~Lopes, Jon Shlens, Ekin~Dogus Cubuk, and Justin
  Gilmer.
\newblock A fourier perspective on model robustness in computer vision.
\newblock In {\em Advances in Neural Information Processing Systems},
  volume~32, 2019.

\bibitem{yosinski2014transferable}
Jason Yosinski, Jeff Clune, Yoshua Bengio, and Hod Lipson.
\newblock How transferable are features in deep neural networks?
\newblock In {\em Advances in Neural Information Processing Systems}, pages
  3320--3328, 2014.

\bibitem{yu2020bdd100k}
Fisher Yu, Haofeng Chen, Xin Wang, Wenqi Xian, Yingying Chen, Fangchen Liu,
  Vashisht Madhavan, and Trevor Darrell.
\newblock Bdd100k: A diverse driving dataset for heterogeneous multitask
  learning.
\newblock In {\em IEEE/CVF Conference on Computer Vision and Pattern
  Recognition}, pages 2636--2645, 2020.

\bibitem{yun2019cutmix}
Sangdoo Yun, Dongyoon Han, Seong~Joon Oh, Sanghyuk Chun, Junsuk Choe, and
  Youngjoon Yoo.
\newblock Cutmix: Regularization strategy to train strong classifiers with
  localizable features.
\newblock In {\em IEEE/CVF International Conference on Computer Vision}, pages
  6023--6032, 2019.

\bibitem{zhang2018mixup}
Hongyi Zhang, Moustapha Cisse, Yann~N Dauphin, and David Lopez-Paz.
\newblock mixup: Beyond empirical risk minimization.
\newblock In {\em International Conference on Learning Representations}, 2018.

\bibitem{zhang2022spectral}
Jingyi Zhang, Jiaxing Huang, Zichen Tian, and Shijian Lu.
\newblock Spectral unsupervised domain adaptation for visual recognition.
\newblock In {\em IEEE/CVF Conference on Computer Vision and Pattern
  Recognition}, pages 9829--9840, 2022.

\bibitem{zhang2018collaborative}
Weichen Zhang, Wanli Ouyang, Wen Li, and Dong Xu.
\newblock Collaborative and adversarial network for unsupervised domain
  adaptation.
\newblock In {\em IEEE/CVF Conference on Computer Vision and Pattern
  Recognition}, pages 3801--3809, 2018.

\bibitem{zhou2020learning}
Kaiyang Zhou, Yongxin Yang, Timothy Hospedales, and Tao Xiang.
\newblock Learning to generate novel domains for domain generalization.
\newblock In {\em European Conference on Computer Vision}, pages 561--578.
  Springer, 2020.

\bibitem{zhou2020domain}
Kaiyang Zhou, Yongxin Yang, Yu Qiao, and Tao Xiang.
\newblock Domain generalization with mixstyle.
\newblock In {\em International Conference on Learning Representations}, 2020.

\end{thebibliography}
}

\end{document}